\documentclass[11pt]{article}

\usepackage[preprint]{acl}

\usepackage{times}
\usepackage{latexsym}
\usepackage{amsmath,amssymb}
\usepackage{subcaption}
\usepackage{booktabs} 
\usepackage{multirow}
\usepackage{xcolor}
\usepackage{colortbl}

\usepackage[T1]{fontenc}

\usepackage[utf8]{inputenc}

\usepackage{microtype}

\usepackage{graphicx}

\title{PA1-OPD: One-Step Prefix-Augmented On-Policy Distillation \\ for Language Model Pretraining}

\author{
  \textbf{Zhenhe Wu\textsuperscript{1}},
  \textbf{Jian Yang\textsuperscript{2}},
  \textbf{Yaping Jin\textsuperscript{1}},
  \textbf{Xinlai Kang\textsuperscript{1}},
  \textbf{Jie Hu\textsuperscript{1}},
  \textbf{Hanting Chen\textsuperscript{1}\footnotemark[1]}
  \\
  \textsuperscript{1}Huawei Technologies,
  \textsuperscript{2}Beihang University
  \\
  \texttt{\{wuzhenhe2,chenhanting\}@huawei.com}
}

\begin{document}
\maketitle
\footnotetext{$^*$Corresponding author.}

\begin{abstract}
Knowledge distillation (KD) improves language-model pretraining by matching teacher distributions at corpus prefixes, but provides no direct supervision at contexts induced by student predictions. We introduce PA1-OPD, which augments corpus-path distillation with teacher feedback after one student-sampled transition per selected prefix. This expands the supervised states while retaining next-token prediction and corpus-path KD. Data-anchored roots support parallel sampling, and differentiable tree attention shares prefix computation without changing inference. In two continued-pretraining settings, full coverage improves held-out cross-entropy over KD across C4, PG-19, and Pile at every evaluated checkpoint. In Stage2, selecting the 10\% of roots with highest student surprisal adds 48.85\% to KD's Macro cross-entropy gain over next-token training. Gains remain when branch supervision excludes successor states already on the corpus path, supporting the value of feedback beyond observed contexts. Additional depth or breadth improves aggregate results at unchanged per-term supervision weight, with benefits varying by corpus. These findings support local student-induced supervision as a complement to pretraining distillation.

\end{abstract}

\section{Introduction}
\label{sec:pa1-introduction}

Knowledge distillation (KD) enriches pretraining with a teacher's predictive distributions beyond observed next-token targets~\cite{hinton2015distilling}. Conventional pretraining KD applies these targets at corpus prefixes. Student predictions can induce different contexts, leaving their successor states without direct supervision. Beyond the choice of distillation loss, this raises the question of which prediction states should receive teacher feedback.

On-policy distillation learns from student-generated trajectories~\cite{ross2011reduction,agarwal2024onpolicy,gu2024minillm}. For pretraining, corpus prefixes provide many roots from which to sample student transitions. Even one sampled token creates a successor state with a new teacher target. This allows us to test whether local feedback beyond the corpus trajectory improves KD without generating complete continuations.

\begin{figure*}[!t]
\centering
\includegraphics[width=\textwidth]{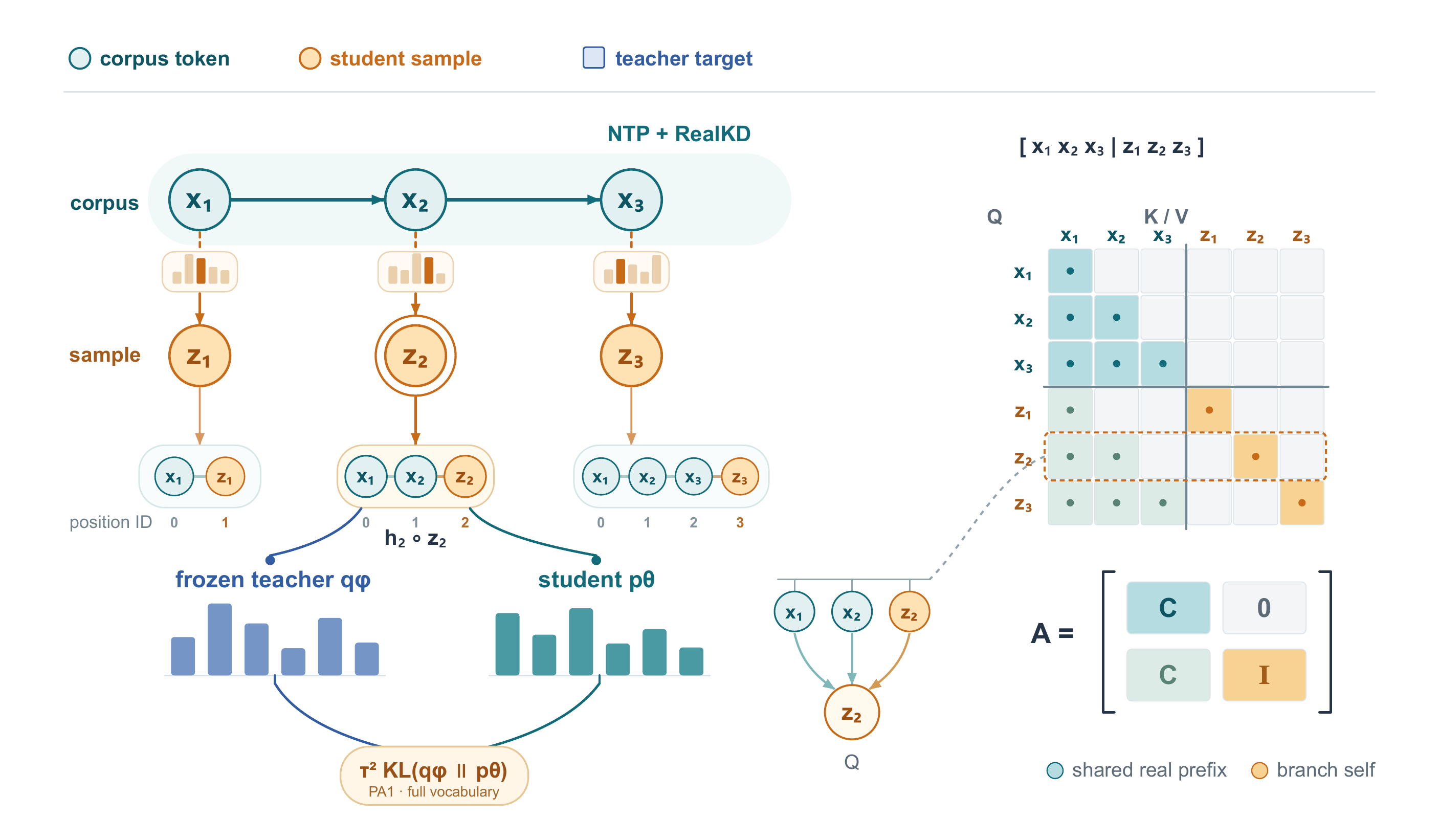}
\caption{PA1-OPD adds teacher supervision after student-sampled transitions from corpus prefixes. Tree attention shares each prefix and isolates branches. Sampling is detached; student prefix states remain differentiable.}
\label{fig:pa1-opd-overview}
\end{figure*}

We propose \textbf{PA1-OPD}, one-step prefix-augmented on-policy distillation (Figure~\ref{fig:pa1-opd-overview}). At each selected corpus prefix, the current student samples one token. Both models consume it, and the teacher supervises the student's next-token distribution at the resulting successor state. This branch objective complements next-token prediction and corpus-path KD. Its on-policy component is local: the transition follows the current student, while its root remains anchored to observed data.

Data-anchored roots support parallel sampling across positions, and tree attention shares their prefixes while keeping branches independent. Differentiable student prefixes retain branch gradients; coincident-state reuse avoids reevaluating corpus states; and chunked KL reduces temporary probability storage. Student-surprisal selection controls where feedback is applied. These operations leave the student architecture and inference unchanged.

We evaluate two OLMo2 continued-pretraining settings. Full coverage improves on KD across all three evaluation corpora in both settings. In Stage2, selecting 10\% of roots adds \textbf{61.60\%, 22.24\%, and 57.12\%} to KD's CE gains over next-token training on C4, PG-19, and Pile (48.85\% Macro). Gains persist without branch losses at coincident corpus states. Stage1 favors full coverage over Top10, while deeper or wider supervision improves Macro CE in both settings. These results connect the gains over KD to the allocation of branch supervision.

\begingroup
\interlinepenalty=10000
\begin{itemize}
\item A pretraining objective that extends teacher supervision to local student-induced states alongside corpus-path training.
\item A differentiable shared-prefix implementation that supports parallel branch supervision and preserves the sampled objective through state reuse and memory-efficient KL computation.
\item Evidence of gains over corpus-path KD, with analyses of noncoincident states, root selection, and additional depth and breadth.
\end{itemize}
\endgroup

\section{Problem Formulation}
\label{sec:pa1-problem}

\subsection{Pretraining Distillation}

We study autoregressive language-model pretraining through checkpoint continuation, using student \(p_\theta\), frozen teacher \(q_\phi\), and corpus \(\mathcal D\). The models share a tokenizer and vocabulary \(\mathcal V\), but may differ in depth and width. Training combines supervision at corpus prefixes and sampled successor states; inference uses only the student.

\subsection{Corpus and Student-Induced States}

For sequence \(x_{b,1:L}\), let \(h_{b,i}=x_{b,1:i}\) denote the prefix through position \(i\), with observed next token \(x_{b,i+1}\). Let \(\mathcal R_b\) be the set of valid next-token prediction positions and \(N_R=\sum_b|\mathcal R_b|\). Standard next-token prediction minimizes

\begin{equation}
\label{eq:pa1-ntp}
\mathcal L_{\mathrm{NTP}}
=-\frac{1}{N_R}
\sum_b\sum_{i\in\mathcal R_b}
\log p_\theta(x_{b,i+1}\mid h_{b,i}).
\end{equation}

Let \(p_\theta^\tau\) and \(q_\phi^\tau\) denote softmax distributions with logits divided by temperature \(\tau>0\). For context \(h\), write \(q_v=q_\phi^\tau(v\mid h)\) and \(p_v=p_\theta^\tau(v\mid h)\). The scaled forward KL is

\begin{equation}
\label{eq:pa1-kl}
d_\tau(q_\phi,p_\theta;h)=\tau^2\sum_{v\in\mathcal V}q_v\log\frac{q_v}{p_v}.
\end{equation}

Distillation on the observed data trajectory, denoted RealKD, is

\begin{equation}
\label{eq:pa1-realkd}
\mathcal L_{\mathrm{RealKD}}
=\frac{1}{N_R}
\sum_b\sum_{i\in\mathcal R_b}
d_\tau(q_\phi,p_\theta;h_{b,i}).
\end{equation}

NTP and RealKD supervise corpus prefixes. To extend these states, let \(\bar\theta\) denote the current student parameters held fixed during sampling. At an eligible corpus prefix, draw

\begin{equation}
\label{eq:pa1-sampling}
\begin{aligned}
z_{b,i}&\sim p_{\bar\theta}^{\tau_s}(\cdot\mid h_{b,i}),\\
\widetilde h_{b,i}&=h_{b,i}\circ z_{b,i}.
\end{aligned}
\end{equation}

Here \(\tau_s\) is the sampling temperature and \(\circ\) denotes token concatenation. If \(\mu_{\mathcal D}\) is the distribution over eligible corpus prefixes, sampling induces the one-step state distribution

\begin{equation}
\label{eq:pa1-state-distribution}
\mu_{\bar\theta}^{(1)}(\widetilde h)
=\mathbb E_{h\sim\mu_{\mathcal D}}\Bigl[
\sum_{z\in\mathcal V}
p_{\bar\theta}^{\tau_s}(z\mid h)\cdot
\mathbf{1}\{\widetilde h=h\circ z\}
\Bigr].
\end{equation}

The transition follows the current temperature-adjusted student policy from a data-anchored root. Its successor may coincide with a corpus state if the sampled token matches the observed next token. Distillation under \(\mu_{\bar\theta}^{(1)}\) adds local student-induced supervision using a single sampled transition.

\section{Method}
\label{sec:pa1-method}

We first define the additional supervision at sampled successor states and its allocation across roots. We then derive a shared-prefix execution that preserves the branch objective and student gradients.

\subsection{One-Step Distillation}
\label{sec:pa1-one-step}

Both models consume the token \(z_{b,i}\) sampled in Eq.~\eqref{eq:pa1-sampling}. The branch loss matches their next-token distributions at the resulting state:

\begin{equation}
\label{eq:pa1-branch-loss}
\ell_{b,i}^{\mathrm{branch}}
=d_\tau(q_\phi,p_\theta;h_{b,i}\circ z_{b,i}).
\end{equation}

The branch target concerns the prediction \emph{after} the sampled token, rather than that token's probability at its root. The observed token \(x_{b,i+1}\) still supplies the NTP target on the corpus path, while the branch receives the teacher's predictive distribution over the full vocabulary.

Let \(\mathcal B_b\subseteq\mathcal R_b\) contain the eligible branch roots and let \(N_B=\sum_b|\mathcal B_b|\). The expected branch objective conditional on a corpus batch is

\begin{equation}
\label{eq:pa1-expected-objective}
\begin{aligned}
\mathcal L_{\mathrm{PA1}}(\theta;\bar\theta)
&=\frac{1}{N_B}
\sum_b\sum_{i\in\mathcal B_b}
\mathbb E_{z\sim p_{\bar\theta}^{\tau_s}(\cdot\mid h_{b,i})}\\
&\qquad\left[
d_\tau(q_\phi,p_\theta;h_{b,i}\circ z)
\right].
\end{aligned}
\end{equation}

One categorical sample from the full vocabulary at each eligible root gives the Monte Carlo estimate

\begin{equation}
\label{eq:pa1-estimator}
\widehat{\mathcal L}_{\mathrm{PA1}}
=\frac{1}{N_B}
\sum_b\sum_{i\in\mathcal B_b}\ell_{b,i}^{\mathrm{branch}}.
\end{equation}

Combining branch and corpus-path losses gives

\begin{equation}
\label{eq:pa1-total-loss}
\boxed{
\mathcal L={}\mathcal L_{\mathrm{NTP}}
+\lambda_{\mathrm{real}}\mathcal L_{\mathrm{RealKD}}
+\lambda_{\mathrm{branch}}\widehat{\mathcal L}_{\mathrm{PA1}}.
}
\end{equation}

The coefficients allocate supervision to corpus and branch states. We differentiate Eq.~\eqref{eq:pa1-total-loss} with the teacher and sampling policy fixed, then refresh the sampling policy after the student update. Gradients do not pass through the sampling distribution~\cite{agarwal2024onpolicy}; they do pass through all student computations used for branch prediction, including the shared prefix.

Eligible roots exclude ignored labels, invalid inputs, and configured document boundaries. NTP and RealKD normalize by \(N_R\), and PA1 by \(N_B\), using optimizer-batch counts to preserve normalization across accumulation steps.

\paragraph{Selecting branch roots.}
Full PA1 uses every eligible root. Selective coverage ranks roots by student surprisal,
\(s_{b,i}=-\log p_\theta(x_{b,i+1}\mid h_{b,i})\),
which is already available from NTP. For retained fraction \(\rho\), let \(\mathcal S_b\) contain the top \(K_b=\lceil\rho|\mathcal B_b|\rceil\) roots. The selected branch loss is
\begin{equation}
\label{eq:pa1-selected-objective}
\widehat{\mathcal L}_{\mathrm{PA1}}^{\mathrm{sel}}
=\frac{1}{N_B}\sum_b\frac{|\mathcal B_b|}{K_b}
\sum_{i\in\mathcal S_b}\ell_{b,i}^{\mathrm{branch}}.
\end{equation}
Sequences without eligible roots are omitted. Scores and selection are detached. This normalization retains the nominal branch-loss scale while concentrating supervision on selected roots. Deterministic top-\(K\) selection targets a different state distribution from Full PA1; it does not provide an unbiased estimate of the full-root objective.

\subsection{Prefix Sharing with Tree Attention}
\label{sec:pa1-prefix-parallel}

Branches share their sequence's corpus prefix but retain independent successor states. Suppressing batch indices, let \(X=(x_1,\ldots,x_L)\) and \(Z=(z_1,\ldots,z_L)\) use layout \([X;Z]\), with one branch slot per corpus position. For causal mask \(C_{ij}=\mathbf{1}[j\le i]\), \(C\in\{0,1\}^{L\times L}\), including the diagonal, tree attention uses

\begin{equation}
\label{eq:pa1-tree-mask}
\boxed{
A=
\begin{bmatrix}
C & 0\\
C & I
\end{bmatrix}
},
\qquad I=I_L.
\end{equation}

Rows index query tokens and columns index key/value tokens; an entry of one permits attention. Each block serves a distinct role:

\begin{itemize}
\item \textbf{Real-to-real, \(C\):} each real token retains its original causal prefix.
\item \textbf{Real-to-branch, \(0\):} branch tokens cannot alter any real-path representation.
\item \textbf{Branch-to-real, \(C\):} branch \(z_i\) reads exactly \(x_{1:i}\), the prefix from which it was sampled.
\item \textbf{Branch-to-branch, \(I\):} each branch can read its own token, while branches from different roots remain isolated.
\end{itemize}

With query, key, and value matrices arranged in the same layout, attention on valid query rows is

\begin{equation}
\label{eq:pa1-masked-attention}
\begin{aligned}
\operatorname{Attn}_A(Q,K,V)
&=\operatorname{softmax}(S_A)V,\\
S_A&=\frac{QK^\top}{\sqrt{d_k}}+M(A),\\
M(A)_{ij}
&=
\begin{cases}
0,& A_{ij}=1,\\
-\infty,& A_{ij}=0.
\end{cases}
\end{aligned}
\end{equation}

Ineligible branch rows are suppressed and contribute zero loss.

Position encodings follow the logical paths. With zero-based position IDs for the real sequence, the augmented IDs are

\begin{equation}
\label{eq:pa1-positions}
\operatorname{pos}([X;Z])=[\,0,1,\ldots,L-1\,;\,1,2,\ldots,L\,].
\end{equation}

Branch \(z_i\) receives the position immediately after \(x_i\), regardless of its storage location. Together with Eq.~\eqref{eq:pa1-tree-mask}, these IDs reproduce the positions of independently extended prefixes.

For fixed samples and deterministic operations, tree execution matches independent extended-prefix evaluation. At each layer, corpus states are unchanged and each branch reads its own state and correctly positioned prefix. Induction over layers gives identical branch representations and outputs. Differentiable sharing preserves the summed parameter gradients up to numerical precision.

\begin{table*}[!t]
\centering
\begingroup
\definecolor{pa1gainred}{RGB}{174,48,48}
\definecolor{pa1lossblue}{RGB}{42,88,153}
\newcommand{\paonegain}[1]{\textcolor{pa1gainred}{#1}}
\newcommand{\paoneloss}[1]{\textcolor{pa1lossblue}{#1}}
\linespread{1}\selectfont
\scriptsize
\setlength{\tabcolsep}{2.6pt}
\renewcommand{\arraystretch}{0.89}
\resizebox{0.82\textwidth}{!}{%
\begin{tabular}{@{}cl*{8}{r}@{}}
\toprule
\multirow{2}{*}{Tokens} & \multicolumn{1}{c}{\multirow{2}{*}{Method}} & \multicolumn{2}{c}{C4} & \multicolumn{2}{c}{PG-19} & \multicolumn{2}{c}{Pile} & \multicolumn{2}{c}{Macro} \\
\cmidrule(lr){3-4}\cmidrule(lr){5-6}\cmidrule(lr){7-8}\cmidrule(l){9-10}
& & CE$\downarrow$ & PPL$\downarrow$ & CE$\downarrow$ & PPL$\downarrow$ & CE$\downarrow$ & PPL$\downarrow$ & CE$\downarrow$ & PPL$\downarrow$ \\
\midrule
\multicolumn{10}{c}{\cellcolor{black!7}\textbf{Stage1 initialization; Stage1 mixture}} \\
\midrule
0 & Initial & 3.018891 & 20.468578 & 2.908693 & 18.332821 & 2.489751 & 12.058270 & 2.805778 & 16.953223 \\
\midrule
2.5B & NTP & 2.967517 & 19.443581 & 2.848100 & 17.254962 & 2.433926 & 11.403570 & 2.749848 & 16.034038 \\
 & KD & 2.963171 & 19.359270 & 2.842240 & 17.154150 & 2.428178 & 11.338206 & 2.744530 & 15.950542 \\
 & Full PA1 & 2.962772 & 19.351543 & 2.841985 & 17.149778 & 2.427837 & 11.334337 & \textbf{2.744198} & \textbf{15.945219} \\
 & \quad\emph{extra/KD} & \paonegain{+9.18\%} & \paonegain{+9.16\%} & \paonegain{+4.35\%} & \paonegain{+4.34\%} & \paonegain{+5.93\%} & \paonegain{+5.92\%} & \paonegain{+6.24\%} & \paonegain{+6.38\%} \\
 & Student Top10 PA1 & 2.963262 & 19.361033 & 2.842728 & 17.162522 & 2.428343 & 11.340076 & 2.744778 & 15.954544 \\
 & \quad\emph{extra/KD} & \paoneloss{-2.09\%} & \paoneloss{-2.09\%} & \paoneloss{-8.33\%} & \paoneloss{-8.30\%} & \paoneloss{-2.87\%} & \paoneloss{-2.86\%} & \paoneloss{-4.66\%} & \paoneloss{-4.79\%} \\
\midrule
5B & NTP & 2.960424 & 19.306153 & 2.840500 & 17.124321 & 2.426554 & 11.319810 & 2.742493 & 15.916761 \\
 & KD & 2.955647 & 19.214145 & 2.834079 & 17.014724 & 2.420120 & 11.247208 & 2.736615 & 15.825359 \\
 & Full PA1 & 2.955310 & 19.207683 & 2.833931 & 17.012201 & 2.419792 & 11.243516 & \textbf{2.736344} & \textbf{15.821133} \\
 & \quad\emph{extra/KD} & \paonegain{+7.03\%} & \paonegain{+7.02\%} & \paonegain{+2.30\%} & \paonegain{+2.30\%} & \paonegain{+5.10\%} & \paonegain{+5.08\%} & \paonegain{+4.61\%} & \paonegain{+4.62\%} \\
 & Student Top10 PA1 & 2.955811 & 19.217295 & 2.834757 & 17.026259 & 2.420289 & 11.249111 & 2.736952 & 15.830888 \\
 & \quad\emph{extra/KD} & \paoneloss{-3.43\%} & \paoneloss{-3.42\%} & \paoneloss{-10.56\%} & \paoneloss{-10.52\%} & \paoneloss{-2.63\%} & \paoneloss{-2.62\%} & \paoneloss{-5.73\%} & \paoneloss{-6.05\%} \\
\midrule
7.5B & NTP & 2.956234 & 19.225430 & 2.835583 & 17.040330 & 2.422717 & 11.276451 & 2.738178 & 15.847404 \\
 & KD & 2.951393 & 19.132596 & 2.829144 & 16.930967 & 2.416162 & 11.202776 & 2.732233 & 15.755446 \\
 & Full PA1 & 2.951049 & 19.126014 & 2.829018 & 16.928827 & 2.415858 & 11.199380 & \textbf{2.731975} & \textbf{15.751407} \\
 & \quad\emph{extra/KD} & \paonegain{+7.11\%} & \paonegain{+7.09\%} & \paonegain{+1.96\%} & \paonegain{+1.96\%} & \paonegain{+4.62\%} & \paonegain{+4.61\%} & \paonegain{+4.34\%} & \paonegain{+4.39\%} \\
 & Student Top10 PA1 & 2.951572 & 19.136004 & 2.829731 & 16.940907 & 2.416383 & 11.205255 & 2.732562 & 15.760722 \\
 & \quad\emph{extra/KD} & \paoneloss{-3.68\%} & \paoneloss{-3.67\%} & \paoneloss{-9.12\%} & \paoneloss{-9.09\%} & \paoneloss{-3.37\%} & \paoneloss{-3.36\%} & \paoneloss{-5.53\%} & \paoneloss{-5.74\%} \\
\midrule
10B & NTP & 2.954363 & 19.189502 & 2.832859 & 16.993983 & 2.420158 & 11.247638 & 2.735794 & 15.810375 \\
 & KD & 2.949240 & 19.091439 & 2.826088 & 16.879301 & 2.413549 & 11.173546 & 2.729626 & 15.714762 \\
 & Full PA1 & 2.948886 & 19.084684 & 2.825901 & 16.876146 & 2.413229 & 11.169976 & \textbf{2.729339} & \textbf{15.710268} \\
 & \quad\emph{extra/KD} & \paonegain{+6.91\%} & \paonegain{+6.89\%} & \paonegain{+2.76\%} & \paonegain{+2.75\%} & \paonegain{+4.84\%} & \paonegain{+4.82\%} & \paonegain{+4.65\%} & \paonegain{+4.70\%} \\
 & Student Top10 PA1 & 2.949337 & 19.093283 & 2.826737 & 16.890257 & 2.413768 & 11.175998 & 2.729947 & 15.719846 \\
 & \quad\emph{extra/KD} & \paoneloss{-1.89\%} & \paoneloss{-1.88\%} & \paoneloss{-9.58\%} & \paoneloss{-9.55\%} & \paoneloss{-3.31\%} & \paoneloss{-3.31\%} & \paoneloss{-5.22\%} & \paoneloss{-5.32\%} \\
\midrule
\multicolumn{10}{c}{\cellcolor{black!7}\textbf{Stage2 initialization; Stage2 mixture}} \\
\midrule
0 & Initial & 2.774223 & 16.026168 & 2.609155 & 13.587569 & 2.213465 & 9.147360 & 2.532281 & 12.920366 \\
\midrule
2.5B & NTP & 2.774907 & 16.037129 & 2.609633 & 13.594061 & 2.213954 & 9.151833 & 2.532831 & 12.927674 \\
 & KD & 2.774035 & 16.023154 & 2.608974 & 13.585100 & 2.213241 & 9.145307 & 2.532083 & 12.917854 \\
 & Full PA1 & 2.773709 & 16.017937 & 2.608825 & 13.583075 & 2.212995 & 9.143058 & 2.531843 & 12.914690 \\
 & \quad\emph{extra/KD} & \paonegain{+37.39\%} & \paonegain{+37.33\%} & \paonegain{+22.61\%} & \paonegain{+22.60\%} & \paonegain{+34.50\%} & \paonegain{+34.46\%} & \paonegain{+32.09\%} & \paonegain{+32.22\%} \\
 & Student Top10 PA1 & 2.773357 & 16.012303 & 2.608787 & 13.582563 & 2.212751 & 9.140827 & \textbf{2.531632} & \textbf{12.911897} \\
 & \quad\emph{extra/KD} & \paonegain{+77.64\%} & \paonegain{+77.65\%} & \paonegain{+28.38\%} & \paonegain{+28.32\%} & \paonegain{+68.72\%} & \paonegain{+68.64\%} & \paonegain{+60.29\%} & \paonegain{+60.65\%} \\
\midrule
5B & NTP & 2.774963 & 16.038031 & 2.609645 & 13.594220 & 2.213974 & 9.152016 & 2.532861 & 12.928089 \\
 & KD & 2.774022 & 16.022954 & 2.608970 & 13.585055 & 2.213233 & 9.145231 & 2.532075 & 12.917747 \\
 & Full PA1 & 2.773704 & 16.017861 & 2.608819 & 13.583000 & 2.212979 & 9.142909 & 2.531834 & 12.914590 \\
 & \quad\emph{extra/KD} & \paonegain{+33.79\%} & \paonegain{+33.78\%} & \paonegain{+22.40\%} & \paonegain{+22.42\%} & \paonegain{+34.23\%} & \paonegain{+34.24\%} & \paonegain{+30.66\%} & \paonegain{+30.53\%} \\
 & Student Top10 PA1 & 2.773378 & 16.012633 & 2.608790 & 13.582610 & 2.212761 & 9.140921 & \textbf{2.531643} & \textbf{12.912054} \\
 & \quad\emph{extra/KD} & \paonegain{+68.44\%} & \paonegain{+68.46\%} & \paonegain{+26.71\%} & \paonegain{+26.68\%} & \paonegain{+63.48\%} & \paonegain{+63.55\%} & \paonegain{+54.96\%} & \paonegain{+55.04\%} \\
\midrule
7.5B & NTP & 2.774990 & 16.038465 & 2.609644 & 13.594215 & 2.213961 & 9.151898 & 2.532865 & 12.928193 \\
 & KD & 2.774025 & 16.022996 & 2.608937 & 13.584609 & 2.213204 & 9.144970 & 2.532055 & 12.917525 \\
 & Full PA1 & 2.773673 & 16.017359 & 2.608789 & 13.582592 & 2.212953 & 9.142675 & 2.531805 & 12.914209 \\
 & \quad\emph{extra/KD} & \paonegain{+36.48\%} & \paonegain{+36.44\%} & \paonegain{+21.07\%} & \paonegain{+21.01\%} & \paonegain{+33.16\%} & \paonegain{+33.12\%} & \paonegain{+30.86\%} & \paonegain{+31.09\%} \\
 & Student Top10 PA1 & 2.773411 & 16.013157 & 2.608777 & 13.582430 & 2.212744 & 9.140760 & \textbf{2.531644} & \textbf{12.912116} \\
 & \quad\emph{extra/KD} & \paonegain{+63.63\%} & \paonegain{+63.60\%} & \paonegain{+22.63\%} & \paonegain{+22.70\%} & \paonegain{+60.77\%} & \paonegain{+60.76\%} & \paonegain{+50.86\%} & \paonegain{+50.70\%} \\
\midrule
10B & NTP & 2.775000 & 16.038629 & 2.609643 & 13.594197 & 2.213954 & 9.151834 & 2.532866 & 12.928220 \\
 & KD & 2.774000 & 16.022603 & 2.608928 & 13.584479 & 2.213189 & 9.144831 & 2.532039 & 12.917304 \\
 & Full PA1 & 2.773673 & 16.017354 & 2.608807 & 13.582831 & 2.212951 & 9.142658 & 2.531810 & 12.914281 \\
 & \quad\emph{extra/KD} & \paonegain{+32.80\%} & \paonegain{+32.75\%} & \paonegain{+16.92\%} & \paonegain{+16.97\%} & \paonegain{+31.11\%} & \paonegain{+31.03\%} & \paonegain{+27.69\%} & \paonegain{+27.70\%} \\
 & Student Top10 PA1 & 2.773385 & 16.012743 & 2.608769 & 13.582324 & 2.212752 & 9.140837 & \textbf{2.531635} & \textbf{12.911968} \\
 & \quad\emph{extra/KD} & \paonegain{+61.60\%} & \paonegain{+61.53\%} & \paonegain{+22.24\%} & \paonegain{+22.18\%} & \paonegain{+57.12\%} & \paonegain{+57.03\%} & \paonegain{+48.85\%} & \paonegain{+48.90\%} \\
\bottomrule
\end{tabular}
}%
\endgroup
\caption{Learning curves at matched training-token budgets. Lower CE and PPL are better; Macro averages each metric over the three datasets. \emph{extra/KD} measures additional improvement beyond KD relative to KD's gain over NTP, not percentage reduction in CE or PPL.}
\label{tab:pa1-main-results}
\end{table*}

\paragraph{Differentiable prefix reuse.}
The student first evaluates the real path for sampling probabilities and retains layer-wise keys and values \(K^X,V^X\). Sampled branches then attend to these caches:

\begin{equation}
\label{eq:pa1-cached-attention}
H_i^Z=\operatorname{Attn}\bigl(Q_i^Z,[K_{1:i}^X;K_i^Z],[V_{1:i}^X;V_i^Z]\bigr).
\end{equation}

The shared caches avoid repeated prefix evaluation and remain differentiable. Branch losses therefore update both new-token computations and the prefix states on which they depend. Teacher and student maintain separate caches; gradients are required only for the student.

\paragraph{Fused teacher evaluation.}
Once student samples are available, the teacher evaluates \([X;Z]\) in one layer traversal under Eq.~\eqref{eq:pa1-tree-mask}. Projection, normalization, and feed-forward operations use the combined layout, while attention enforces each branch's context. Fusion reduces repeated traversals and kernel dispatches without removing the computation required by the added states.

Block-sparse attention, native key/value-head counts, and chunked branch queries limit intermediate storage while preserving this graph.

\subsection{Objective-Preserving Execution}
\label{sec:pa1-efficient-execution}

When a sampled token matches the next corpus token, we reuse the available successor logits with student gradients intact and evaluate only unmatched branches. Chunked KL limits temporary probability storage with full-vocabulary normalization. Both preserve the sampled objective: reuse retains coincident-state losses, and chunking retains the complete teacher distribution. Appendix~\ref{sec:pa1-execution-details} details the execution and reuse conditions.

\section{Experiments}
\label{sec:pa1-experiments}

\subsection{Experimental Setup}
\label{sec:pa1-setup}

\paragraph{Models and data.}
We distill a frozen OLMo2-7B teacher into OLMo2-1B~\cite{olmo2024furious}. Students start from two official \texttt{OLMo-2-0425-1B} checkpoints: \texttt{stage1-step30000-tokens63B} (\textbf{Stage1}) and the Stage2-final \texttt{main} revision (\textbf{Stage2}). Stage1 uses the official Stage1 mixture; Stage2 uses Dolmino Stage2. Each continues for approximately 10B tokens with a fresh optimizer, sharing data and order across methods within each setting. Evaluation uses fixed C4 validation, PG-19 test, and Pile validation data~\cite{raffel2020exploring,rae2019compressive,gao2020pile}; Appendix~\ref{sec:pa1-training-details} specifies the full protocol.

\paragraph{Comparisons.}
We compare NTP, corpus-path KD (\(\mathcal L_{\mathrm{NTP}}+0.7\mathcal L_{\mathrm{RealKD}}\)), and PA1-OPD (\(\mathcal L_{\mathrm{NTP}}+0.5\mathcal L_{\mathrm{RealKD}}+0.2\widehat{\mathcal L}_{\mathrm{PA1}}\)). PA1 reallocates part of the nominal distillation weight to branch states. Full PA1 uses every eligible root; Student Top10 retains the 10\% with highest student surprisal. Both temperatures are 1. Each condition uses one run, with training tokens matched within its setting.

\paragraph{Metrics.}
We report token-averaged cross-entropy (CE, in nats) and dataset perplexity, \(\mathrm{PPL}=\exp(\mathrm{CE})\). Macro CE and Macro PPL separately average the three dataset scores. To quantify the additional distillation benefit when KD improves on NTP, define
\begin{equation}
\label{eq:pa1-extra-kd}
\emph{extra/KD}
=100\%\frac{L_{\mathrm{KD}}-L_{\mathrm{PA1}}}
{L_{\mathrm{NTP}}-L_{\mathrm{KD}}}.
\end{equation}
Here \(L\) is the stated metric; Macro ratios use Macro scores rather than averaged dataset ratios. This ratio measures improvement beyond KD relative to KD's gain over NTP, not relative reduction in CE or PPL. Absolute scores accompany the ratios because small denominators can magnify them. All subsequent analyses use CE.

\paragraph{Analysis protocol.}
Controlled analyses use 1B-token continuations from Stage2, and also from Stage1 for branch expansion. Positional diagnostics use fixed training sequences at checkpoints from the main Top10 runs. These analyses are distinct from the main 10B-token comparison; Appendix~\ref{sec:pa1-analysis-details} gives their experimental configurations.

\begin{figure}[!t]
\centering
\includegraphics[width=\columnwidth]{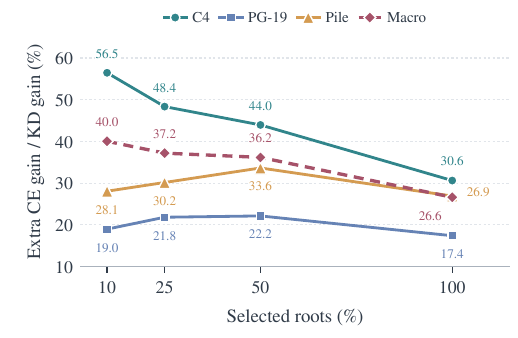}
\caption{Stage2 root-coverage comparison after 1B continuation tokens with student-surprisal selection. Curves report CE \emph{extra/KD}; higher is better.}
\label{fig:pa1-selection-budget}
\end{figure}

\begin{figure*}[!t]
\centering
\captionsetup[subfigure]{font=footnotesize,skip=2pt,justification=centering}
\begin{subfigure}[t]{0.98\textwidth}
\centering
\includegraphics[width=\linewidth]{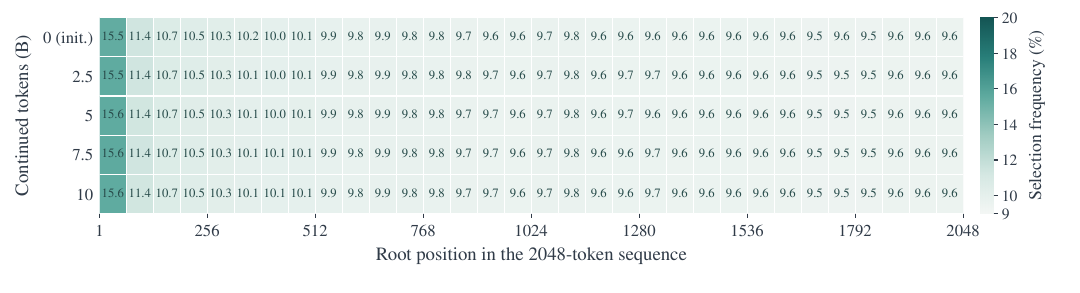}
\caption{Stage1 checkpoint}
\label{fig:pa1-positions-63b}
\end{subfigure}
\par\vspace{4pt}
\begin{subfigure}[t]{0.98\textwidth}
\centering
\includegraphics[width=\linewidth]{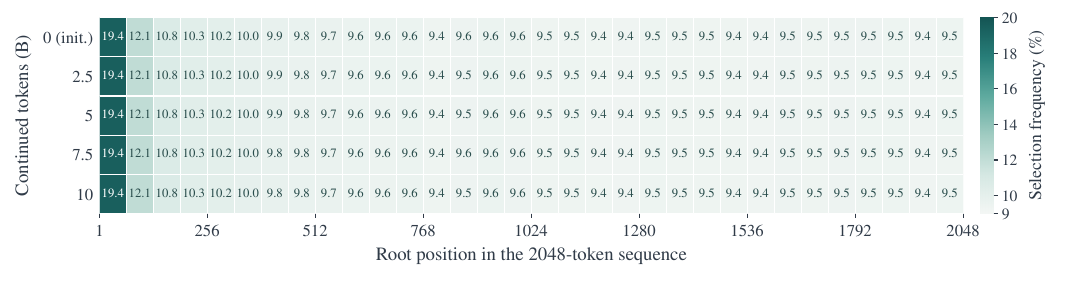}
\caption{Stage2 checkpoint}
\label{fig:pa1-positions-stage2}
\end{subfigure}
\caption{Positional distribution of selected roots. Cell values give selection frequencies among eligible roots in each position bin, with a shared color scale across settings and checkpoints.}
\label{fig:pa1-position-selection}
\end{figure*}

\subsection{Main Results}
\label{sec:pa1-main-results}

\paragraph{Local supervision adds to corpus-path KD.}
Full PA1 improves CE and PPL over KD on all three datasets at every evaluated checkpoint in both settings (Table~\ref{tab:pa1-main-results}). After 10B tokens, its Macro CE is lower by \(2.87\times10^{-4}\) in Stage1 and \(2.29\times10^{-4}\) in Stage2, adding 4.65\% and 27.69\% to KD's gain over NTP. The direction of improvement holds across the reported training trajectory and does not depend on a single evaluation corpus or checkpoint.

\paragraph{Coverage controls the additional benefit.}
Student Top10 improves on Full PA1 and KD throughout Stage2. Its final Macro CE is 2.531635 versus 2.532039 for KD, an absolute reduction of \(4.04\times10^{-4}\) and 48.85\% \emph{extra/KD}. Stage2 NTP ends above its initial CE, so the ratio compares continuation objectives rather than gains over initialization. In Stage1, Top10 underperforms KD at every checkpoint, ending at \(-5.22\%\) \emph{extra/KD}, whereas Full PA1 remains beneficial. Thus, local branch supervision transfers across both tested settings, but aggressive root selection does not. Initialization and mixture change together, so coverage should be chosen for the continuation setting rather than inferred from checkpoint stage alone.

\paragraph{Additional state evaluations.}
For a length-\(L\) sequence \(b\), Top10 selects \(K=\lceil0.1|\mathcal B_b|\rceil\) roots. Before reuse, its additional forward token-state evaluations satisfy
\begin{equation}
\label{eq:pa1-forward-states}
\Delta N_{\mathrm{fwd}}\le 2K\le 2\lceil0.1L\rceil\approx0.2L.
\end{equation}
Each model therefore adds at most approximately 10\% to its \(L\) corpus states before reuse; KD already evaluates \(2L\) states across both models. Reuse can reduce this increment further. These state counts describe the added supervision workload. They exclude model-size differences, attention, and backward computation, and cannot establish gains in compute efficiency.

\section{Analysis}
\label{sec:pa1-discussion}

We examine three aspects of the added supervision: whether gains require coincident corpus states, how root selection distributes feedback, and whether deeper or wider signals add further benefit. Comparisons follow the controlled protocols in Section~\ref{sec:pa1-setup} and Appendix~\ref{sec:pa1-analysis-details}.

\begin{table}[!t]
\centering
\begingroup
\small
\setlength{\tabcolsep}{3pt}
\begin{tabular}{@{}lrrrr@{}}
\toprule
Variant & C4 & PG-19 & Pile & Macro \\
\midrule
Full PA1 & 30.63 & 17.41 & 26.91 & 26.63 \\
Teacher Top50 & 33.52 & 19.30 & 28.77 & 29.07 \\
Student Top50 & 43.96 & 22.15 & 33.64 & 36.18 \\
Drop coincident & 44.37 & 26.58 & 34.80 & 37.80 \\
\bottomrule
\end{tabular}
\endgroup
\caption{Stage2 root-selection and coincident-state ablations as CE \emph{extra/KD} percentages (higher is better). \emph{Drop coincident} uses Student Top50 and removes only losses on states already present on the corpus path.}
\label{tab:pa1-selection-overlap}
\end{table}

\subsection{Supervision Beyond Corpus States}
\label{sec:pa1-overlap-supervision}

A branch can revisit a corpus-path state when its sampled token matches the observed continuation. To test whether these repeated targets are necessary, we remove only the corresponding branch KL terms while retaining the original normalization. The resulting variant still improves on KD on every dataset (Table~\ref{tab:pa1-selection-overlap}). Its Macro CE is 2.531165, compared with 2.531173 when coincident terms are retained. The small difference does not establish an advantage for removal; the substantive result is that feedback at noncoincident student-induced states remains useful. Loss removal changes supervision; state reuse instead preserves all loss terms while avoiding redundant state evaluations.

\subsection{Allocating Feedback Across Roots}
\label{sec:pa1-root-budget}
\label{sec:pa1-root-scoring}

Sparse selection improves the controlled Stage2 results (Figure~\ref{fig:pa1-selection-budget}): Macro \emph{extra/KD} increases from 26.63\% with full coverage to 40.04\% with Top10. Every tested sparse variant improves on Full PA1 for each dataset. C4 favors the smallest tested budget, whereas PG-19 and Pile favor broader coverage. Equation~\eqref{eq:pa1-selected-objective} retains nominal branch weight while concentrating it on fewer roots, so coverage controls both the number and distribution of supervised states. Stage1 favors broader coverage, indicating that concentrated feedback is useful in some continuation settings but not others.

At matched 50\% coverage, student surprisal outperforms \emph{teacher advantage}, the teacher's observed-token probability minus the student's. Macro \emph{extra/KD} rises from 29.07\% to 36.18\% (Table~\ref{tab:pa1-selection-overlap}). Available directly from NTP, surprisal selects roots where the observed continuation is difficult for the student and improves on the tested teacher-based scoring rule.

\begin{figure*}[!t]
\centering
\captionsetup[subfigure]{font=footnotesize,skip=2pt,justification=centering}
\begin{subfigure}[t]{0.48\textwidth}
\centering
\includegraphics[width=\linewidth]{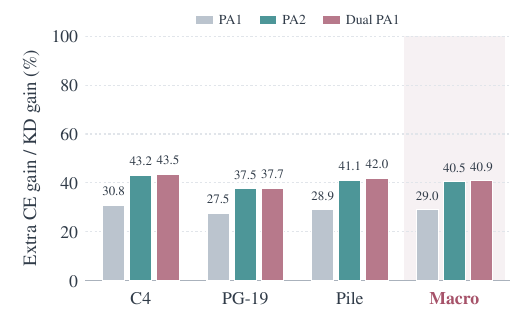}
\caption{Stage1 checkpoint}
\label{fig:pa1-branch-expansion-63b}
\end{subfigure}
\hfill
\begin{subfigure}[t]{0.48\textwidth}
\centering
\includegraphics[width=\linewidth]{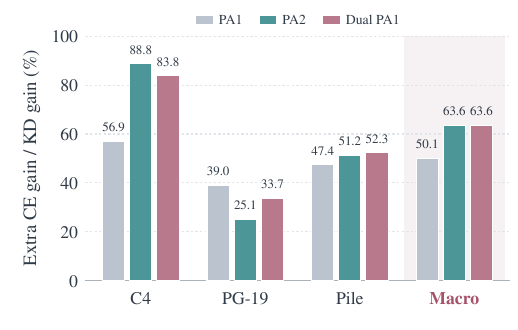}
\caption{Stage2 checkpoint}
\label{fig:pa1-branch-expansion-stage2}
\end{subfigure}
\caption{Full-root branch expansion after 1B continuation tokens. Each branch term retains weight 0.2. CE \emph{extra/KD} uses matched NTP and KD references with KD weight 0.5; higher is better.}
\label{fig:pa1-branch-expansion}
\end{figure*}

\subsection{Coverage Across Context Lengths}
\label{sec:pa1-position-selection}
\label{sec:pa1-position-details}

We next examine whether surprisal selection concentrates supervision on short contexts. Figure~\ref{fig:pa1-position-selection} reports frequencies in 64-position bins, recomputed at each checkpoint on the same 8,192 training sequences within each setting. The profiles therefore reflect the current selector rather than accumulated training-time choices. At the final checkpoint, the first bin has selection rates of 15.57\% in Stage1 and 19.42\% in Stage2, yet accounts for only 4.86\% and 6.06\% of selected roots. Latter-half selection rates remain 9.61\% and 9.47\%, close to the overall Top10 rate. Early-position preference thus coexists with broad coverage of longer prefixes.

The largest bin-wise changes from initialization are 0.122 and 0.043 percentage points. These small changes show stable aggregate positional preferences across the continuation, even though the individual selected roots may change with the student.

\subsection{Adding Depth and Breadth}
\label{sec:pa1-branch-expansion}

We also test whether adding supervision at greater depth or breadth improves on PA1. \textbf{PA\(x\)} follows \(x\) sequential student samples from each corpus prefix and applies teacher feedback after every transition. PA2 adds a second step; \emph{dual PA1} instead draws two independent one-step samples with replacement from the same root. Each branch KL term retains weight 0.2, and corpus-path KD retains weight 0.5. The extensions therefore add a second supervision signal at unchanged per-term strength, increasing total nominal branch weight from 0.2 to 0.4.

After 1B continuation tokens, both extensions improve on PA1 across all three Stage1 corpora (Figure~\ref{fig:pa1-branch-expansion}). PA2 and dual PA1 reduce Macro CE by \(4.38\times10^{-4}\) and \(4.56\times10^{-4}\), raising \emph{extra/KD} from 28.99\% to 40.47\% and 40.94\%, respectively. Their Macro CE difference is only \(1.8\times10^{-5}\). In Stage2, both lower Macro CE by approximately \(5.6\times10^{-5}\), raising \emph{extra/KD} from 50.12\% to 63.61\%. C4 and Pile improve, while PG-19 favors single PA1. Additional supervision therefore benefits the aggregate metric in both settings, with corpus-dependent returns and no established general preference for depth over breadth.

Both extensions double sampled tokens and supervised branch states per root before reuse; PA2 also requires sequential sampling. They test added feedback at fixed per-term strength. Together with root selection, these results show that the amount and placement of branch supervision both affect the gains from local student-induced states.

\paragraph{Implications for pretraining.}
The comparisons separate three allocations of teacher feedback. The main study shifts part of a fixed nominal KD weight to student-induced successors. Root selection concentrates that weight on fewer successors; depth and breadth add another KL signal at fixed per-term strength. Gains with noncoincident states support local state augmentation as the core method. Selection and expansion then control its distribution and amount, with benefits that vary across the evaluated continuation settings.

\section{Related Work}
\label{sec:pa1-related-work}

\paragraph{Distillation and student-induced supervision.}
KD transfers predictive distributions, sequences, and representations~\cite{hinton2015distilling,kim2016sequence,sanh2019distilbert,jiao2020tinybert}; pretraining applications study compression, data refinement, and compute allocation~\cite{gemma2024gemma2,muralidharan2024compact,gu2025miniplm,busbridge2025distillation}. DAgger and scheduled sampling motivate learning at learner-induced states~\cite{ross2011reduction,bengio2015scheduled}, and ImitKD, GKD, and MiniLLM apply related principles to distillation~\cite{lin2020autoregressive,agarwal2024onpolicy,gu2024minillm}. Subsequent work explores objectives and data reuse~\cite{ko2024distillm,ko2025distillm2,jin2026entropy}, interleaved sampling~\cite{xu2025speculative}, and token or rollout-prefix selection~\cite{huang2025speckd,zhang2026fast,zhao2026prefix}. Broader efforts explore task-specific supervision and context construction~\citep{wu-etal-2026-table,wu2026mrsql,wu-etal-2025-ucs}, alongside cache-aware expert management for MoE inference~\citep{wu2026cacheaware}. PA1-OPD focuses on a complementary design choice: applying forward-KL supervision after local student transitions from corpus prefixes during pretraining.

\paragraph{Additional computation and execution.}
Repeated tokens, pause tokens, and multi-token prediction add computation or prediction targets~\cite{wu2025efficient,goyal2024think,gloeckle2024better}. PA1-OPD adds teacher-supervised training states without changing inference, adapting shared-prefix trees from speculative verification~\cite{miao2024specinfer,cai2024medusa} for differentiable student computation. Cut Cross-Entropy avoids global logit materialization~\cite{wijmans2025cut}; our chunked KL retains logits but reduces probability workspace, preserving the teacher distribution.

\section{Conclusion}
\label{sec:pa1-conclusion}

PA1-OPD augments pretraining KD with teacher supervision after local student-generated transitions. Full coverage improves on corpus-path KD in both evaluated continuation settings, and Stage2 Top10 adds 48.85\% to KD's Macro CE gain over NTP. Gains remain without coincident-state branch losses, supporting feedback beyond observed corpus states. Additional depth or breadth further improves aggregate results at fixed per-term weight, while coverage and corpus determine the returns. Teacher-supervised state allocation thus offers a practical design dimension for improving pretraining distillation with local student feedback.

\clearpage
\section*{Limitations}

Our evaluation covers one teacher--student family, two continuation settings, and three held-out corpora. Each condition uses one training run, so independent-seed replication is needed to assess the small absolute improvements. The results establish observed likelihood gains in these settings, not generality across model families, scales, languages, or downstream tasks. 

The main KD and PA1 objectives also allocate their nominal distillation weights differently; additional controls could separate weighting effects from supervision location. Our experiments cover PA1 and its extensions to two sequential steps (PA2) or two branches per root (dual PA1), with fixed per-term branch weights. Future work could extend PA1 to PA\(x\) with \(x>2\) and explore wider branching to examine how supervision depth and breadth affect pretraining gains. Comparisons match training tokens rather than compute, and state counts do not measure runtime or FLOP efficiency. These boundaries leave open how the observed gains scale with longer pretraining and additional teacher feedback.

\section*{Ethical Considerations}

This study uses publicly available language models and existing pretraining and evaluation corpora, with no human participants or collection of new personal data. Distillation may inherit biases or harmful content from the teacher and training data. Our likelihood evaluations do not assess these effects, and no method-specific ethical risks were identified in the reported experiments.

\bibliography{custom}%

\clearpage
\appendix

\section{Execution and Cost Analysis}
\label{sec:pa1-execution-details}

We expand the execution in Section~\ref{sec:pa1-efficient-execution} with a tree-mask example, reuse conditions, KL gradients, and computational cost accounting.

\paragraph{Three-root attention example.}
For three real positions, the structure is explicit:

\begin{equation}
\label{eq:pa1-mask-example}
\begin{array}{c|ccc|ccc}
 & x_1 & x_2 & x_3 & z_1 & z_2 & z_3\\ \hline
x_1 & 1 & 0 & 0 & 0 & 0 & 0\\
x_2 & 1 & 1 & 0 & 0 & 0 & 0\\
x_3 & 1 & 1 & 1 & 0 & 0 & 0\\ \hline
z_1 & 1 & 0 & 0 & 1 & 0 & 0\\
z_2 & 1 & 1 & 0 & 0 & 1 & 0\\
z_3 & 1 & 1 & 1 & 0 & 0 & 1
\end{array}.
\end{equation}

The three branches represent \((x_1,z_1)\), \((x_1,x_2,z_2)\), and \((x_1,x_2,x_3,z_3)\), respectively. The identity block prevents cross-branch communication despite their joint tensor layout.

\paragraph{Reusing states on the data trajectory.}
When \(z_i=x_{i+1}\) and the next input is available, the branch matches corpus state \(h_{i+1}\). Deterministic evaluation with identical positions gives

\begin{equation}
\label{eq:pa1-coincident-paths}
\begin{aligned}
p_\theta(\cdot\mid h_i\circ z_i)
&=p_\theta(\cdot\mid h_{i+1}),\\
q_\phi(\cdot\mid h_i\circ z_i)
&=q_\phi(\cdot\mid h_{i+1}).
\end{aligned}
\end{equation}

After checking input validity and document boundaries, we gather each model's real-path logits and compact unmatched branches. Samples, branch losses, and \(N_B\) remain unchanged. Differentiable student gathers retain both the RealKD and preceding root's branch terms, preserving losses and gradients up to numerical precision. When all branches coincide, existing corpus logits supply every branch prediction.

\paragraph{Memory-efficient full-vocabulary KL.}
We evaluate KL in vocabulary chunks with full-vocabulary normalization. For logits \(a\) and vocabulary partition \(\{\mathcal V_c\}\), the log-normalizer is

\begin{equation}
\label{eq:pa1-log-normalizer}
\log Z_a
=\operatorname{logaddexp}_c
\left[
\operatorname{logsumexp}_{v\in\mathcal V_c}(a_v/\tau)
\right].
\end{equation}

We accumulate Eq.~\eqref{eq:pa1-kl} across the same chunks in FP32. With a frozen teacher, the derivative with respect to the student logit \(a_v\) is

\begin{equation}
\label{eq:pa1-kl-gradient}
\frac{\partial d_\tau}{\partial a_v}
=\tau\left(p_\theta^\tau(v)-q_\phi^\tau(v)\right).
\end{equation}

The backward pass saves logits and per-position scalars and recomputes chunk probabilities. It retains teacher mass \(m_T\) (one in exact arithmetic) for numerical consistency, computing \(\tau(p_\theta^\tau(v)m_T-q_\phi^\tau(v))\).

For \(P\) positions and chunk width \(c\), the FP32 workspace is \(O(Pc)\), with \(O(P)\) saved scalars. Full logits and student-logit gradients remain stored. Chunking retains the complete teacher distribution and temperature scaling.

\paragraph{Training procedure and computational cost.}
For corpus batch \(X\), student \(\theta\), and frozen teacher \(\phi\), a PA1-OPD update consists of:
\begin{enumerate}
\item Determine valid real prediction positions and eligible branch roots, and count them over the complete optimizer batch.
\item For each accumulation microbatch:
    \begin{enumerate}
    \item Evaluate the student real path and retain differentiable K/V caches.
    \item Select roots as needed and sample one token per root from detached logits.
    \item Identify branches that coincide with available states on the corpus path.
    \item Evaluate unmatched student branches using shared real-prefix K/V.
    \item Evaluate teacher real and unmatched branch paths with the tree mask.
    \item Gather real-path logits for matched states and align branch outputs.
    \item Compute NTP, RealKD, and PA1 using complete-batch denominators, and accumulate student gradients.
    \end{enumerate}
\item Aggregate gradients across workers and update the student parameters.
\end{enumerate}

Ignoring padding, each model evaluates \(L+U\) states for a sequence of length \(L\) and \(U\) unmatched branches. Projection and feed-forward work scale with the state count, while prefix computation is shared. For unmatched roots \(\mathcal U\) with one-based prefix lengths \(i\), the allowed attention pairs total

\begin{equation}
\label{eq:pa1-attention-cost}
N_{\mathrm{attn}}
=\frac{L(L+1)}{2}+\sum_{i\in\mathcal U}(i+1).
\end{equation}

The terms count real-path causal pairs and branch attention to prefixes and themselves. Branches add \(O(UL)\) pairs without cross-root dependencies. Block granularity and padding affect kernel work but preserve this graph. All added computation is confined to training.

\section{Experimental Details}
\label{sec:pa1-training-details}

\paragraph{Backbone models.}
The frozen OLMo~2 teacher~\cite{olmo2024furious} is \texttt{OLMo-2-1124-7B} (32 layers, hidden size 4096). Both students share the \texttt{OLMo-2-0425-1B} architecture (16 layers, hidden size 2048) and Dolma2 tokenizer. Section~\ref{sec:pa1-setup} defines the initializations; Stage2 uses commit \texttt{a1847dff}. Continuations load student weights with a fresh optimizer.

\paragraph{Datasets.}
For Stage1, we draw approximately 10B tokens according to the official token-weighted proportions of its seven-component mixture, including configured source repeats (chiefly DCLM, with code, scientific, mathematical, and Wikipedia text). For Stage2, we follow the six-component proportions of its 50B-token Dolmino-Mix-1124 recipe (DCLM, math, FLAN, Wikipedia, PeS2o, and StackExchange). In each setting, sampling selects nonoverlapping contiguous blocks aligned to 2048-token sequences; all methods share the data and order. Each run consumes exactly 10,001,317,888 training tokens. We evaluate on fixed tokenized files from C4 validation (web text)~\cite{raffel2020exploring}, PG-19 test (long-form books)~\cite{rae2019compressive}, and Pile validation (multi-domain text)~\cite{gao2020pile}. All complete 2048-token windows are scored, yielding 9.998M, 9.871M, and 9.998M evaluation tokens, respectively.

\paragraph{Implementation details.}
The main experiments are conducted on a single training node equipped with eight accelerator cards. Training uses BF16 autocasting, 2048-token sequences, one sequence per GPU per microbatch, and a global batch of 2,097,152 tokens (1024 sequences). Each run takes 4,769 optimizer steps, with checkpoints at approximately 2.5B, 5B, 7.5B, and 10B tokens. AdamW uses \((\beta_1,\beta_2)=(0.9,0.95)\), gradient clipping at 1, and weight decay 0.033 in Stage1 or 0.1 in Stage2; embeddings are exempt from decay. Stage1 uses an initial learning rate of \(10^{-4}\) with its inherited cosine schedule (zero warmup, final ratio 0.1, \(T_{\max}=1{,}907{,}359\) steps); Stage2 uses a constant \(3\times10^{-6}\). Branch attention selects FlexAttention when available, with 256-query reference chunks otherwise; full-vocabulary KL uses chunks of 8192. Model and data seeds are 12536 and 34521, respectively. Evaluation uses batch size four. 

\subsection{Analysis Protocols}
\label{sec:pa1-analysis-details}

Unless stated otherwise, controlled comparisons continue the Stage2 student for 1B tokens using the same teacher as the main experiments. They use a fixed Dolmino Stage2 sample, 477 updates, 2048-token sequences, a global batch of 2,097,152 tokens, and a constant learning rate of \(3\times10^{-6}\). Four- and eight-GPU runs preserve the global batch through gradient accumulation. Branch expansion also uses the Stage1 checkpoint.

\paragraph{Root coverage and scoring.}
Student Top10, Top25, and Top50 retain the indicated percentage of eligible roots in each sequence using student surprisal and the normalization in Eq.~\eqref{eq:pa1-selected-objective}. Teacher Top50 instead ranks the same candidates by
\(q_\phi(x_{b,i+1}\mid h_{b,i})-p_\theta(x_{b,i+1}\mid h_{b,i})\).
All use the same nominal PA1 objective weights.

\paragraph{Coincident-state supervision.}
The removal ablation retains Student Top50 selection but zeros branch KL terms for sampled successors that coincide with available corpus states. NTP, RealKD, and the original branch denominator remain unchanged. Both variants use coincident-state reuse.

\paragraph{Positional diagnostics.}
For each main Top10 run, we sample 8,192 training sequences uniformly without replacement and hold them fixed across initialization and the 2.5B, 5B, 7.5B, and 10B checkpoints. Selection is recomputed before state reuse. We pool selected and eligible occurrences into 64-position bins and divide selected counts by eligible counts; ineligible roots, including position 2048, are excluded. These frequencies describe each checkpoint's selector; they do not aggregate selections across training.

\paragraph{Branch expansion.}
The full-root comparisons in Figure~\ref{fig:pa1-branch-expansion} use 1B-token continuations (477 updates). The Stage1 runs share the historical DCLM-based Stage1-like data and order, learning rate \(10^{-4}\), global batch 2,097,152, sequence length 2048, and historical optimizer settings and schedule. Both new extensions use eight GPUs and coincident-state reuse. Stage2 follows the protocol above. Each panel uses NTP and KD references from the same checkpoint and continuation budget, with KD weight 0.5. Macro ratios follow Eq.~\eqref{eq:pa1-extra-kd}. Scores use six decimal places; the displayed Stage2 tie need not imply identical unrounded CE.

\end{document}